\documentclass{article}
\usepackage{spconf,amsmath,graphicx}

\usepackage{booktabs}
\usepackage{siunitx}
\usepackage{caption}
\usepackage{graphicx}
\usepackage{amssymb}
\usepackage{comment}
\usepackage{amssymb}
\usepackage{hyperref}
\usepackage[all=normal,charwidths=tight,leading=tight,
]{savetrees}
\title{CLIP-guided Multi-task Regression for Multi-View Plant Phenotyping}
\name{Simon Warmers$^{\dagger}$, Muhammad Zawish$^{\ddagger}$, Fayaz Ali Dharejo$^{\dagger}$, Steven Davy$^{\ddagger}$, Radu Timofte$^{\dagger}$\thanks{This work was supported by the Alexander von Humboldt Foundation.}}
\address{$^{\dagger}$Computer Vision Lab, CAIDAS, IFI, University of W\"urzburg, Germany \\
$^{\ddagger}$Technological University Dublin, Ireland}

\begin{document}
%
\maketitle
\begin{abstract}
Modeling plant growth dynamics plays a central role in modern agricultural research. However, learning robust predictors from multi-view plant imagery remains challenging due to strong viewpoint redundancy and viewpoint-dependent appearance changes. We propose a level-aware vision language framework that jointly predicts plant age and leaf count using a single multi-task model built on CLIP embeddings. Our method aggregates rotational views into angle-invariant representations and conditions visual features on lightweight text priors encoding viewpoint level for stable prediction under incomplete or unordered inputs. On the GroMo25 benchmark, our approach reduces mean age MAE from 7.74 to 3.91 and mean leaf-count MAE from 5.52 to 3.08 compared to the GroMo baseline, corresponding to improvements of 49.5\% and 44.2\%, respectively. The unified formulation simplifies the pipeline by replacing the conventional dual-model setup while improving robustness to missing views. The modela and code is available at: https://github.com/SimonWarmers/CLIP-MVP 
\end{abstract}
\section{Introduction}
\label{sec:intro}
Plant phenotyping from multiview imagery is crucial for precision agriculture, enabling non-invasive monitoring of growth traits such as age and leaf count \cite{yang2025deep, campillo2010study}. Large-scale, multi-view plant phenotyping benchmarks exemplified by the GroMo25 challenge \cite{bhatt2025gromo} have catalyzed rapid progress in automated plant age and leaf-count estimation, but they also expose a fundamental modeling tension: how to aggregate hundreds of highly redundant, viewpoint-correlated images into compact, robust predictions without losing growth-stage signal or overfitting to view-specific artifacts. Prior challenge baselines \cite{bansal2025gromo25} and multiview transformers \cite{bhatt2025gromo, dosovitskiy2020image} demonstrate that incorporating many views can improve accuracy, yet they rely on heavy, task-specific architectures or explicit view-selection heuristics to tame redundancy.

Recent work has attacked the redundancy problem from the data-selection and sampling side. For example, ViewSparsifier \cite{kampa2025viewsparsifier} shows that carefully choosing a sparse subset of views can recover performance while drastically reducing computation. However, these approaches still treat view management and downstream trait estimation as separate problems, and they do not exploit cross-task feature sharing between correlated phenotypic targets. In practice, (i)\textit{ multi-model pipelines increase cost and error propagation between components}, and (ii) \textit{deployed systems must often operate with incomplete or user-captured view sets}, which breaks methods that assume dense, well-ordered multi-view inputs.

In this work, rather than building bespoke models per trait or relying on sparse view selection as a preprocessing crutch, we design a single, unified vision–language model that jointly predicts plant age and leaf count from multiview inputs while explicitly reasoning about viewpoint level and input completeness. Our approach builds on CLIP \cite{chen2022contrastive}, a large-scale vision–language model trained on image–text pairs that has shown strong generalization across diverse visual concepts \cite{du2024teach}, and which we reformulate from classification to regression by topping its multimodal representations with a lightweight MLP regressor.

Concretely, our level-aware multimodal CLIP embedding pipeline aggregates rotational views into angle-invariant level representations, conditions these visual codes on lightweight CLIP text embeddings that encode height and level priors, and trains a multi-task regressor end-to-end so that morphological cues useful for one trait can inform the other. Crucially, the textual conditioning acts as a flexible run-time guide. When some views or metadata are missing, the model predicts or retrieves an appropriate text embedding through a learned level regressor, resolving viewpoint ambiguity instead of failing outright.

We summarize our contributions as follows. (1) We introduce a single-model, multi-task framework that replaces the conventional dual-model paradigm for age and leaf-count estimation, enabling positive transfer across traits and streamlining inference. (2) We propose a level-aware multimodal fusion strategy that combines CLIP visual embeddings (augmented by Grounding DINO preprocessing \cite{liu2024grounding}) with compact CLIP text priors to disentangle viewpoint-induced appearance changes from genuine phenotypic variation, and we show how a learned level estimator supplies this guidance at test time when metadata are absent. (3) We demonstrate that this multimodal conditioning yields consistent gains and greater robustness to incomplete multi-view inputs on the GroMo25 benchmark \cite{bhatt2025gromo, bansal2025gromo25} : our level-aware model reduces mean age MAE from 4.12 to 3.91 and mean leaf-count MAE from 3.43 to 3.08 relative to a strong unimodal CLIP baseline, while markedly reducing performance degradation as views are removed.

\section{Related Work}
\label{sec:rw}
Leaf counting and plant growth estimation have been active research areas in plant phenotyping. Traditional deep learning approaches explore direct regression and segmentation-assisted pipelines \cite{abe2024promptable, zawish2024energy, aich2017leaf, bhagat2022eff}. For example, Authors in \cite{farjon2021leaf} investigated both direct and detection-augmented regression for leaf traits [6], while Dobrescu et al. \cite{dobrescu2017leveraging} showed that deep models trained on CVPPP \cite{scharr2017computer} emphasize plant structure over background during counting \cite{dobrescu2019understanding}. Other works have used architectures ranging from lightweight detectors such as Tiny-YOLOv3 \cite{deb2024cnn} to segmentation-based pipelines with deconvolutional and encoder–decoder backbones. More recent designs combine hierarchical features and spatial detail through two-stream and pyramid structures \cite{fan2022segmentation}. These models have been evaluated on datasets such as Arabidopsis \cite{buzzy2020real}, CVPPP \cite{scharr2017computer}, but they primarily focus on single-view imagery captured from fixed overhead angles, limiting their ability to reason about 3D structure or handle occlusions from complex canopy geometries.

The recently introduced GroMo25 dataset \cite{bansal2025gromo25, bhatt2025gromo} and challenge formalize a multi-view phenotyping benchmark with two core tasks: plant age prediction and leaf count estimation using 24 rotational views at five height levels. This dataset exposes several challenges: high redundancy across views, strong viewpoint correlations, and the need to integrate vertical and horizontal cues to capture growth dynamics efficiently. Recent methods like ViewSparsifier \cite{kampa2025viewsparsifier} have shown that randomly selecting and aggregating view subsets can reduce redundancy and improve view-invariant embedding learning, winning both GroMo tasks by learning compact multi-view features that generalize across crops and views. Despite this progress, existing GroMo25 solutions \cite{bansal2025gromo25} treat age and leaf count as separate objectives and rely on view selection heuristics rather than integrated multi-task reasoning. To our knowledge, no prior work has explored a unified multi-view, multi-task model that jointly predicts age and leaf count while leveraging multimodal priors to handle missing views or metadata. Our work addresses this gap with a level-aware vision–language model that fuses visual evidence and textual guidance for robust, joint phenotypic estimation.

\section{Methodology}
\subsection{Overview}
This section presents the proposed unified vision–language pipelines for joint prediction of plant age and leaf count from multiview image data. The goal is to replace the dual‐model paradigm of prior work \cite{bhatt2025gromo, bansal2025gromo25, kampa2025viewsparsifier} with a single, shared model that jointly learns both traits from a common visual representation.  We first describe shared preprocessing and representation steps, and then introduce three successive pipeline variants. Our design is motivated by the hypothesis that vision-language conditioning can guide the regression head to resolve ambiguities in plant appearance (e.g. due to viewpoint variation). Figure 2 illustrates the overall progression starting with a purely visual baseline to architectures that introduce explicit linguistic conditioning and self-descriptive feedback.

\begin{figure}[!t]
    \centering
    \includegraphics[width=\linewidth, keepaspectratio]{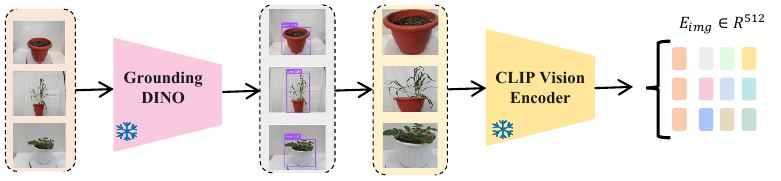}
   \caption{The proposed preprocessing pipeline based on Grounding DINO \cite{liu2024grounding} and CLIP \cite{chen2022contrastive} that encodes images into rich visual embeddings for downstream tasks.}
   \label{fig:real_syn_pred}  
\end{figure}

The GroMo 2025 dataset poses a unique challenge due to its multi-view acquisition protocol, with plants captured from arbitrary angles, varying heights, and heterogeneous backgrounds. This leads to view-correlated redundancy and spatially inconsistent background noise, which degrades the discriminative power of downstream vision encoders. At inference time, uncontrolled user framing and environmental clutter further amplify the domain shift between training and deployment. To address this, we employ a two-stage preprocessing pipeline (Figure \ref{fig:real_syn_pred}). First, a pretrained Grounding DINO \cite{liu2024grounding} model performs object-centric localization to generate tight bounding boxes around the plant, enabling adaptive cropping that preserves semantically relevant regions (plant and pot) while suppressing background. The crops are then resized and encoded by a CLIP visual encoder into 512-dimensional embeddings that are robust to view and scale variations. These embeddings are cached for training, avoiding repeated preprocessing and yielding cleaner, more consistent visual representations.

\begin{figure*}[t]
\centering
\includegraphics[width=0.7\textwidth]{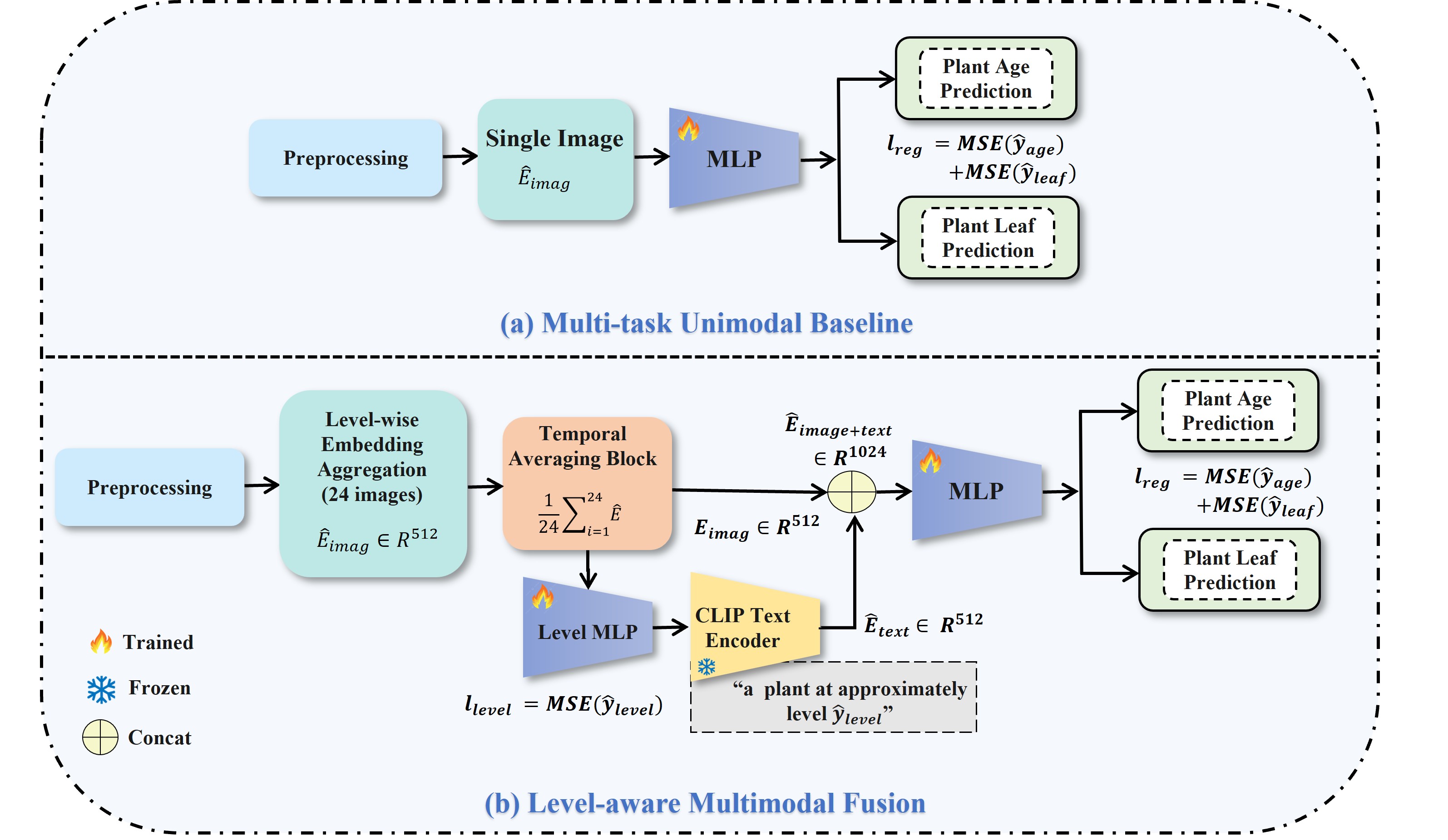}
\caption{The proposed multi-task pipelines: (a) shows unimodal baseline based on CLIP vision embeddings, (b) shows conditioning CLIP on level priors for multimodal regression.}
\label{fig1}
\end{figure*}

\subsection{Multi-task Unimodal Baseline}
The original approach proposed for GroMo dataset adopt a single-task learning strategy, where distinct models are trained independently for each phenotypic attribute (i.e, plant age and leaf count). This approach neglects the inherent correlations between these traits and results in redundant computation, and limited feature transferability across related prediction tasks. To overcome these shortcomings, we propose a multi-task baseline that jointly predicts plant age and leaf count from a unified representation space, thereby exploiting shared morphological and textural information.

The proposed approach builds upon the CLIP image embeddings as the input feature space thanks to its strong generalization capabilities and rich semantic priors. CLIP, pretrained on hundreds of millions of image and text pairs, captures high-level abstractions such as object structure, spatial context, and compositional semantics. These representations have demonstrated robustness across diverse visual domains, making them particularly suitable for downstream tasks in agricultural imaging where annotated data are often scarce. After standard preprocessing, each input image is encoded using CLIP's vision encoder, yielding a 512-dimensional embedding vector 
$\hat{E}_{\text{imag}} \in \mathbb{R}^{512}$. 
This embedding is then passed through a lightweight fully connected MLP, designed with an input layer matching the embedding dimension and an output layer of size two, corresponding to the regression targets: plant age and leaf count. The model is trained end-to-end using a composite multi-task loss that simultaneously optimizes both outputs, enabling implicit feature sharing and improved generalization across traits. Despite its simplicity, this baseline establishes a strong foundation, demonstrating the efficacy of pretrained multimodal representations for multi-trait plant phenotyping.

\begin{table*}[!ht]
\centering
\footnotesize
\setlength{\tabcolsep}{5pt}
\renewcommand{\arraystretch}{1.2}
\caption{Comparison of different approaches with proposed methods based on Mean Absolute Error (MAE) for plant age and leaf count prediction. Lower values indicate better performance.}
\label{tab:approach_comparison}
\begin{tabular}{lcccccccc}
\toprule
{\textbf{Approach}} & \multicolumn{4}{c}{\textbf{MAE (Age)}} & \multicolumn{4}{c}{\textbf{MAE (Leaf Count)}} \\
\cmidrule(lr){2-5} \cmidrule(lr){6-9}
& \textbf{Mustard} & \textbf{Radish} & \textbf{Wheat} & \textbf{Mean} & \textbf{Mustard} & \textbf{Radish} & \textbf{Wheat} & \textbf{Mean} \\
\midrule
GroMo~[8] & 10.62 & 5.71 & 8.80 & 7.74 & 4.99 & 4.04 & 10.80 & 5.52 \\
\midrule
CropIQ \cite{bansal2025gromo25} & 21.70 & 16.54 & 28.60 & 19.41 & 5.65 & 6.48 & 8.70 & 5.83 \\
Agro\_Geek \cite{bansal2025gromo25} & 11.30 & 18.85 & 28.45 & 18.01 & 7.38 & 3.98 & 30.41 & 10.90 \\
Rishi \cite{bansal2025gromo25} & 12.64 & 10.08 & 16.79 & 12.98 & 4.15 & 6.89 & 5.18 & 8.68 \\
SoumikDas \cite{bansal2025gromo25} & 10.18 & 2.68 & 10.60 & 8.65 & 3.33 & 1.33 & 7.20 & 3.73 \\
PlantPixels \cite{bansal2025gromo25} & 3.20 & 5.60 & 7.30 & 7.30 & 8.30 & 0.66 & 5.68 & 4.35 \\
AlgriTech \cite{bansal2025gromo25} & 8.70 & 5.03 & 8.44 & 6.48 & 4.40 & 1.79 & 6.71 & 3.63 \\
DeepLeaf \cite{bansal2025gromo25} & 7.80 & 4.60 & 6.15 & 5.83 & 7.60 & 0.89 & 5.25 & 3.69 \\
ViewSparsifier \cite{kampa2025viewsparsifier} & 6.47 & 1.61 & 2.90 & 3.66 & 2.70 & 0.83 & 3.38 & 2.30 \\
\midrule
Proposed Unimodal (Image only) & 4.24 & 2.68 & 5.41 & 4.12 & 4.69 & 2.45 & 3.16 & 3.43 \\
Proposed Level-aware Multimodal & 4.48 & 2.44 & 4.80 & 3.91 & 4.81 & 1.19 & 3.23 & 3.08 \\
\bottomrule
\end{tabular}
\end{table*}

\begin{table}[t]
\centering
\caption{Comparison of multimodality and multitask capability across different approaches.}
\label{tab:multimodal_multitask}
\setlength{\tabcolsep}{4pt} 
\begin{tabular}{lcc}
\hline
\textbf{Approach} & \textbf{Multimodal} & \textbf{Multitask} \\
\hline
GroMo~\cite{bhatt2025gromo} & $\times$ & $\times$ \\
CropIQ~\cite{bansal2025gromo25} & $\times$ & $\times$ \\
Agro Geek~\cite{bansal2025gromo25} & $\times$ & $\times$ \\
Rishi~\cite{bansal2025gromo25} & $\times$ & $\times$ \\
SoumikDas~\cite{bansal2025gromo25} & $\times$ & $\times$ \\
PlantPixels~\cite{bansal2025gromo25} & $\times$ & $\times$ \\
AlgriTech~\cite{bansal2025gromo25} & $\times$ & $\times$ \\
DeepLeaf~\cite{bansal2025gromo25} & $\times$ & $\times$ \\
ViewSparsifier~\cite{kampa2025viewsparsifier} & $\times$ & $\times$ \\
\hline
Proposed Unimodal (Image only) & $\times$ & $\checkmark$ \\
Proposed Level-aware Multimodal & $\checkmark$ & $\checkmark$ \\
\hline
\end{tabular}
\end{table}

\subsection{Level-aware Multimodal Fusion}
The baseline approach presented above processes each image independently, similar to the original approach by GroMo \cite{bhatt2025gromo}. Such methods ignore the structured spatial design of the GroMo25 dataset, which contains recordings from five height levels and 24 rotational views per level per day. This design provides dense coverage but also introduces redundancy. Many neighbouring views look almost identical, while significant changes appear only after several rotations. More importantly, it creates a semantic ambiguity: plant appearance changes both with growth stage and viewpoint height. Without height information, the model cannot separate these two factors. For example, a young plant seen from a low angle may resemble the base of a mature plant. A dense canopy from above may look like an older plant, even if it is simply compact. This confusion leads to unstable predictions across unseen or incomplete sets of images.

To address this, we introduce explicit height-level conditioning. Each image is first encoded using CLIP's vision encoder to obtain a 512-dimensional embedding, $\hat{E}^{(i)}_{\text{img}} \in \mathbb{R}^{512}$. For each level, we aggregate information from the 24 rotational views by computing their element-wise mean, i.e., $\bar{E}_{\text{level}} = \frac{1}{24} \sum_{i=1}^{24} \hat{E}^{(i)}_{\text{img}}$, which results in a single, angle-invariant representation $\bar{E}_{\text{level}} \in \mathbb{R}^{512}$.
The averaging operation reduces redundancy from highly similar adjacent views and improves robustness when some views are missing or occluded. Crucially, it produces a fixed-dimensional output, allowing the model to handle variable input completeness without retraining.

During training, each consolidated visual embedding $\bar{E}_{\text{level}}$ is paired with its known height level. We use CLIP's text encoder with the prompt ``\textit{a plant at approximately level X}'', where $X \in \{1,\dots,5\}$. This produces a text embedding $\hat{E}_{\text{text}} \in \mathbb{R}^{512}$ that encodes geometric context. During inference, the level may not be known. To handle this, an auxiliary regressor predicts the most likely height level $\hat{\ell}$ from the visual embedding. The predicted level $\hat{\ell}$ is then used to generate the text embedding, providing contextual guidance even when metadata are missing.

The visual and text embeddings are concatenated into a 1024-dimensional vector, $E_{\text{fused}} = [\bar{E}_{\text{level}} \, \| \, \hat{E}_{\text{text}}] \in \mathbb{R}^{1024}$, and passed through an MLP to predict plant age and leaf count. Encoding height information allows the network to interpret similar visual cues differently depending on the viewpoint. For example, a dense cluster of leaves indicates canopy maturity at Level 5 but suggests overlapping lower leaves at Level 2. The model therefore learns to associate visual features with their correct geometric context, effectively disentangling viewpoint artifacts from genuine phenotypic traits. This level-aware representation significantly improves robustness, generalization, and interpretability, especially when only partial multi-view data are available.

\section{Experiments and Results}
\subsection{Dataset Preparation}
All experiments were conducted using the GroMo 2025 Challenge dataset \cite{bhatt2025gromo}, a large-scale, multi-view plant phenotyping benchmark. Each plant is imaged from 24 distinct viewpoints at five height levels, resulting in a total of 120 images per plant per day. In accordance with the official challenge guidelines, each height level is processed independently; that is, only the 24 views corresponding to a single height level are used for training and evaluation at that level.

\subsubsection{Cleaning} The dataset contains several inconsistencies due to its size and heterogeneous collection process. Erroneous or missing entries were manually removed or corrected in the CSV files, which map image filenames to plant metadata such as age and leaf count. Each species has four CSV files covering all plants; we focus on mustard, radish, and wheat. The mustard subset had the most errors, including missing images (e.g., plants 1 and 2 on day 12) and mismatched filenames. In wheat, the fifth height level had only 23 valid images instead of 24 and was excluded from experiments 6 and 7. Leaf count annotations include all visible leaves, even from other plants. For instance, mustard plant 1 on day 11 shows a jump from 14 to 40 leaves due to weeds, which were removed later. Following the dataset authors’ protocol, such cases were retained, as they had negligible impact on model convergence and performance.

\subsubsection{Preprocessing}
A custom data loading pipeline was built to handle preprocessing and error checking. During preprocessing, Grounding DINO \cite{liu2024grounding} was used to find where the plant is located in each image. Instead of using fixed class labels, Grounding DINO takes text prompts and links them to matching visual regions. In this work, it helped detect and crop the part of the image containing only the plant and its pot. This step removed most of the background. The cropped images were then resized to match the input size required by the pretrained CLIP model \cite{chen2022contrastive}. During the generation of CLIP embeddings, corrupted or mismatched samples were identified and corrected within the CSV files. The final data loader performed an additional validation step to skip any remaining inconsistencies, thereby ensuring a clean and stable input stream throughout training.

\subsection{Implementation}

\subsubsection{Unimodal Baseline}
For the unimodal baseline, the regression model is a small MLP with ReLU activations and two regression heads implemented as a shared final linear layer with two outputs. The input to the model is a 512-dimensional image embedding. The hidden layers have sizes 1024, 512, and 64, each followed by a ReLU activation. The output layer consists of two units representing the predicted \textit{age} and \textit{leaf count}. The loss function is defined as the sum of the mean squared errors (MSE) of the two regression targets, i.e., $L = \mathrm{MSE}(\text{age}) + \mathrm{MSE}(\text{leaf\_count})$. The model is optimized using the Adam optimizer with a learning rate of 0.001, a batch size of 64, and trained for 10 epochs. All training was performed on an NVIDIA GeForce RTX 3060.

\subsubsection{Multimodal Fusion}
For the multimodal setting, an MLP with ReLU activations was used to combine image and text embeddings. The model input is a 1024-dimensional vector formed by concatenating a 512-dimensional image embedding with a 512-dimensional CLIP text embedding. The text embedding encodes the prompt \textit{``plant at approximately level $\hat{y}_{\text{level}}$''}. The value $\hat{y}_{\text{level}}$ is obtained from a separate MLP discussed in Section 1 of Supplementary. The multimodal MLP consists of hidden layers of sizes 2048, 1024, 512, and 64, each followed by a ReLU activation. The output layer is a shared linear layer with two outputs corresponding to \textit{age} and \textit{leaf count}. The same loss function is used as in the unimodal case. The model is trained using the Adam optimizer with a learning rate of 0.001, a batch size of 64, and for 10 epochs. To make the training more efficient, the 24 image embeddings corresponding to one height level are averaged to form a single 512-dimensional image representation before concatenation. The text embeddings are L2-normalized prior to fusion. To further reduce computation time, the CLIP text embeddings for all five levels are precomputed and stored in a lookup table, which replaces direct CLIP inference during training.

\begin{figure}[t]
\centering
\includegraphics[width=0.5\textwidth]{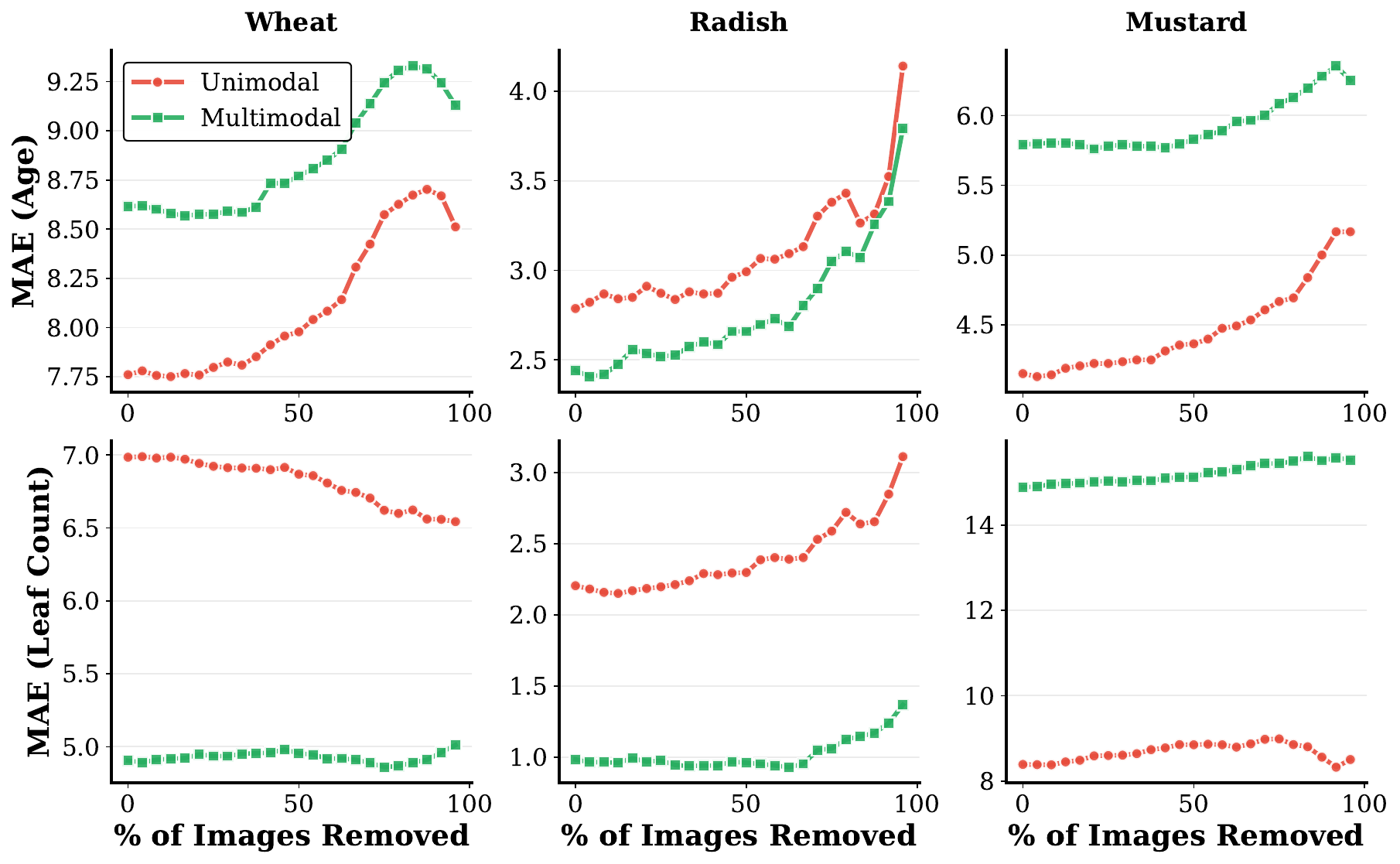}
\caption{MAE of both approaches as a function of the percentage of viewpoint images removed at runtime.}
\label{fig3}
\end{figure}

\subsection{Evaluation}
We evaluate both the unimodal baseline and the proposed level-aware multimodal approach using MAE, the official metric of the GroMo 2025 challenge. To ensure a realistic assessment of generalization, we adopt a plant-based data split: for each species, one plant is held out for testing and the remaining plants are used for training. This \textit{plant split} avoids leakage caused by similar viewpoints or height levels across images and ensures evaluation on entirely unseen plants, yielding a more reliable measure of real-world performance. Table~1 reports a comparison between our methods, the GroMo baseline, and state-of-the-art approaches. On the GroMo25 benchmark, our approach reduces the mean age MAE from 7.74 to 3.91 and the mean leaf-count MAE from 5.52 to 3.08 compared to the GroMo baseline, corresponding to improvements of 49.5\% and 44.2\%, respectively. Our method outperforms all competing approaches except ViewSparsifier \cite{kampa2025viewsparsifier}; however, as shown in Table~2, our approach is more efficient in practice, since we propose a single multimodal multitask model, whereas ViewSparsifier requires a separate model for each task. Overall, our approach achieves a strong balance between accuracy and efficiency, delivering state-of-the-art performance with a more compact and unified modeling framework.

\begin{figure}[t]
\centering
\includegraphics[width=0.3\textwidth]{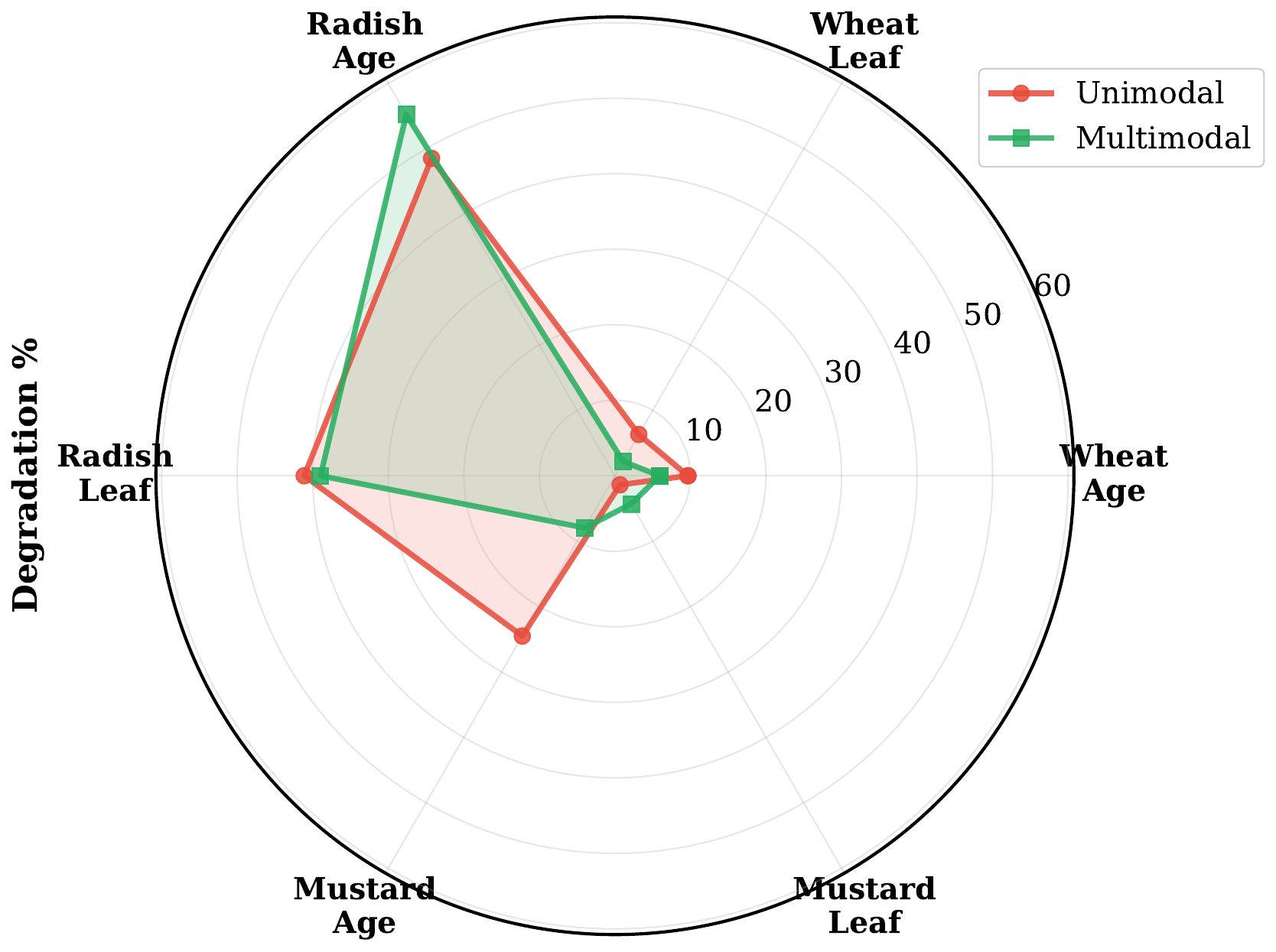}
\caption{Overall degradation from 0\% removal to only 1 image remaining.}
\label{fig4}
\end{figure}
\subsubsection{Sensitivity to Missing Viewpoints}
Figure~3 shows the MAE of both approaches as a function of the percentage of viewpoint images removed at runtime. Each plant is imaged from 24 distinct viewpoints, but in practice, some viewpoints may be missing or occluded. To test robustness, we progressively remove images during inference. For age estimation, MAE rises gradually and steepens after 50\% of images are removed, whereas leaf-count predictions remain stable until 70--80\% image removal, demonstrating good resilience of both approaches under partial observation. We analyze overall robustness in Figure~4, where degradation is computed as $ \text{Degradation (\%)} = \frac{\text{Final MAE} - \text{Initial MAE}}{\text{Initial MAE}} \times 100$, with \textit{Initial MAE} measured using all 24 images (0\% removed) and \textit{Final MAE} measured with only 1 image (95.8\% removed). Overall, multimodal shows lower average degradation (19.10\%) compared to unimodal (21.93\%), indicating they are 12.9\% more robust under extreme image removal.

\section{Conclusion}
We introduce a multimodal regression method utilizing CLIP for plant age and leaf-count estimation that fuses multiview images into angle-invariant representations and leverages CLIP text embeddings for height-level guidance. Our single-model, multi-task approach handles missing viewpoints, reduces redundancy, and enables positive transfer across traits. On GroMo25, it lowers mean age MAE from 4.12 to 3.91 and leaf-count MAE from 3.43 to 3.08, while improving robustness by 12.9\% under extreme view removal. Experiments on mustard, radish, and wheat show that multimodal conditioning boosts age estimation, stabilizes leaf-count predictions, and tolerates dataset inconsistencies. Future work includes extending CLIP-based priors to more traits, dynamic viewpoint selection, and larger, heterogeneous datasets.

\vfill\pagebreak

\bibliographystyle{IEEEbib}
\bibliography{strings,refs}

\end{document}